\documentclass{article}

\usepackage{fullpage}
\usepackage[utf8]{inputenc} 
\usepackage[T1]{fontenc}    
\usepackage{hyperref}       
\usepackage{url}            
\usepackage{booktabs}       
\usepackage{amsfonts}       
\usepackage{nicefrac}       
\usepackage{microtype}      
\usepackage{tikz}
\usepackage{graphicx}
\usepackage{subcaption}
\usepackage[numbered]{algorithm}
\usepackage{algorithmic}
\usepackage{amsmath}
\usepackage[square,numbers]{natbib}
\DeclareMathOperator*{\argmin}{arg\,min}

\title{Critic Algorithms using Cooperative Networks}

\author{Debangshu Banerjee 
\and 
Kavita Wagh}

\begin{document}

\maketitle

\begin{abstract}
  An algorithm is proposed for policy evaluation in Markov Decision Processes which gives good empirical results with respect to convergence rates. The algorithm tracks the Projected Bellman Error and is implemented as a true gradient based algorithm. In this respect this algorithm differs from TD($\lambda$) class of algorithms. This algorithm tracks the Projected Bellman Algorithm and is therefore different from the class of residual algorithms. Further the convergence of this algorithm is empirically much faster than GTD2 class of algorithms which aim at tracking the Projected Bellman Error. We implemented proposed algorithm in DQN and DDPG framework and found that our algorithm achieves comparable results in both of these experiments.  
\end{abstract}

\section{Introduction}

While most reinforcement learning algorithms aim at minimizing the Mean Squared Bellman Error, in function approximation it makes more sense to track the Projected Bellman Error. This is because with function approximation the true optimal of the Bellman Equation might not be representable by the function class. An example would be the true solution not being within the range space of the design matrix when using linear architectures. In such a scenario, one looks at the projected optimal solution onto the range space of the design matrix. This projected optimal solution is the fixed point solution of the Bellman Equation.

\section{Problem Setup}

Let $(\mathcal{X,U,P,R, \pi})$ denote a Markov Decision Problem, where states can take values in a state space $\mathcal{X}$. Corresponding to a state $x \in \mathcal{X}$, we can take an action $u \in \mathcal{U}$, corresponding to which we get a reward $r \in \mathcal{R}$. The actions can be taken according to a strategy or a policy $\mu : \mathcal{X} \to \mathcal{U}$, where corresponding to each state we take an action as mentioned by the policy $\mu$. 

The planning problem or the evaluation problem corresponds to evaluating the goodness of a policy $\mu$. Described formally, being at a state $i$, the evaluation of the goodness of the state while following a strategy $\mu$ is the cumulative expected reward earned $$J_\mu(i) = \mathbf{E}_\mu[\sum_{t=0}^{\infty} g(i_{t}, i_{t+1})| i_0 = i] $$ 

These expectations can be computed as Monte Carlo runs on simulations and can be updated online as a type of Robbins Munro Equation.

From this perspective, let $(i_0, i_1, i_2, ...)$ be a sequence of states as observed while following a policy $\mu$. At any state $i_k$ we can compute the cost-to-go $J_\mu(i_k)$ as follows :
$$J_\mu(i_k) = \mathbf{E}[\sum_{t=0}^{\infty} g(i_{k+t}, i_{k+t+1})] $$
$$ = \mathbf{E}[ g(i_{k}, i_{k+1}) + J_\mu(i_{k+1})] $$
$$ = (1-\lambda)\mathbf{E}[\sum_{l=0}^{\infty}\lambda^l(\sum_{n=0}^{l} g(i_{k+n}, i_{k+n+1}) + J_\mu(i_{k+l+1}))] $$, where $\lambda$ is a parameter or eligibility trace between $(0,1)$ which weights the $l-step$ rewards.

$$ = \mathbf{E}[\sum_{l=k}^{\infty}\lambda^{l-k}d_l] + J_\mu(i_{k}) $$, where $d_l = g(i_l , i_{l+1} + J_\mu(i_{l+1}) - J_\mu(i_l)$

The Robbins Munro stochastic approximation algorithm for the above equations becomes
$$J(i_k) := J(i_k) + \gamma \sum_{l=k}^{\infty}\lambda^{l-k}d_l$$

This form of update is what is known as a look-up table approach, since we need to keep a separate entry for each states' cost to go, and update every entry as the simulation proceeds.

In contrast when it comes to function approximation, we approximate $J_\mu(i)$ by $\tilde{J}(i,r)$ where $r$ is a learnable parameter.

We solve the following minimization problem for instance in the stochastic shortest path problem, where $N$ is a stopping time :
$$\min_r \sum_{k=0}^{N-1}[\sum_{t=k}^{N-1}g(i_{t},i_{t+1}) - \tilde{J}(i_k,r)]^2 $$

Gradient descent gives us the following update
$$r := r - \gamma\sum_{k=0}^{N-1}\nabla\tilde{J}(i_k,r)(\tilde{J}(i_k,r) - \sum_{t=k}^{N-1}g(i_{t},i_{t+1}))$$
$$= r + \gamma\sum_{k=0}^{N-1}\nabla\tilde{J}(i_k,r)(\sum_{t=k}^{N-1}d_t)$$

By allowing for eligibility trace $\lambda$, the update is

$$r := r + \gamma\sum_{k=0}^{N-1}\nabla\tilde{J}(i_k,r)(\sum_{t=k}^{N-1}d_t\lambda^{t-k})$$

And for discounted problems as $$r := r + \gamma\sum_{k=0}^{\infty}\nabla\tilde{J}(i_k,r)(\sum_{t=k}^{\infty}d_t(\alpha\lambda)^{t-k})$$

This can be algabraically rearranged so that we can perform the updates in an online fashion 
$$r := r + \gamma d_k\sum_{t=0}^{k}(\alpha\lambda)^{k-t}\nabla\tilde{J}(i_t,r)$$

For linear architectures, $\tilde{J}(i,r) = \phi(i)^Tr$, where $\phi(i)$ is a design vector, whose dimension $k << |\mathcal{X}|$, the above update is 
$$r_{t+1} = r_t + \gamma_t d_t\sum_{k=0}^{t}(\alpha\lambda)^{t-k}\phi(i_k)$$

It is known the above update, under mild conditions, converge to the fixed point solution of $$\Pi T^\lambda(\Phi r^*) = \Phi r^*$$
where $T^\lambda J(i) = (1-\lambda)\mathbf{E}[\sum_{l=0}^{\infty}\lambda^l(\sum_{n=0}^{l} g(i_{n}, i_{n+1}) + J_\mu(i_{l+1}))| i_0 = i]$ is the Bellman Operator, \\
$\Pi = \Phi (\Phi^TD\Phi)^{-1}\Phi^TD $ is the projection operator on the range space of $\Phi$, \\$\Phi$ is the design matrix $s.t. \tilde{J}(r) = \Phi r$ and $\Phi \in \mathbf{R}^{|\mathcal{X}| \times k}$
and \\ $D = diag\{\pi(i),\pi(j)..\}$, $D \in \mathbf{R}^{|\mathcal{X}| \times |\mathcal{X}|} $ where $\pi$ is the steady state distribution of the Markov Decision Process.

Based on the above equation one has the following Value Iteration update
$$J := \Pi T^\lambda(\Phi r) $$

Though, these algorithms have strong convergence results, a fundamental problem that so exists is that these are not true gradient based methods. To illustrate note that for $\lambda = 0$, get the familiar TD(0) update:
$$r_{t+1} = r_t + \gamma_t d_t\phi(i_t) =  r_t + \gamma_t(g(i_t , i_{t+1}) + \phi(i_{t+1})^Tr_t - \phi(i_t)^Tr_t)\phi(i_t)$$
 A true gradient descent algorithm based on the the following objective function for a 1-step Bellman Error $$\min_r ||\tilde{J}(i,r) - T \tilde{J}(i,r))||_D^2$$ where $T \tilde{J}(i,r)) = \mathbf{E}_\mu[g(i_{t}, i_{t+1}) + \tilde{J}(i_{t+1},r)| i_t = i]$, gives the following update
 $$r_{t+1} = r_t + \gamma_t(g(i_t , i_{t+1}) + \phi(i_{t+1})^Tr_t - \phi(i_t)^Tr_t)(\phi(i_t) - \phi(i_{t+1})) $$
 
 The other issue is, it seems more intuitive to look at the following objective function
 $$\min_r ||\tilde{J}(i,r) - \Pi T \tilde{J}(i,r))||_D^2,$$ where $\Pi$, $T \tilde{J}(i,r))$ are as defined before. We shall call this the Mean Square Projected Bellman Error.

\subsection{Derivation}

We aim at minimizing the following objective function $$\min ||\Pi TJ - J||_D^2$$

where $\Pi = \Phi (\Phi^TD\Phi)^{-1}\Phi^TD$ is the projection operator on the range space of $\Phi$\\ $TJ(i) = \mathbf{E}_\mu[g(i,j) + J(j)]$ is the 1-step Bellman Operator and \\ $J = \Phi r$ where $r \in \mathbf{R}^k$

This objective function was introduced by Sutton et al and is known as the Mean Squared Projected Bellman Error.

A rough idea that we propose here would be to modify the following algorithm 
\begin{algorithm}

\hspace*{\algorithmicindent} \textbf{Input} $J_k$\\
\hspace*{\algorithmicindent} \textbf{Output} $J^*$
\begin{algorithmic}
\REPEAT
\WHILE{$\delta$ tolerance}
\STATE compute $\Pi TJ_k$
\ENDWHILE
\STATE $J_{k+1} \leftarrow \Pi TJ_k $
\UNTIL{$||\Pi TJ_k - J_k||_D^2 < \epsilon$}
\end{algorithmic}
\end{algorithm}
which is not truly gradient based because of 
\begin{algorithmic}
\STATE $J_{k+1} \leftarrow \Pi TJ_k $ 
\end{algorithmic}
to 
\begin{algorithmic}
\WHILE{$\delta$ tolerance}
\STATE  $\min_{J_{k+1}}||\Pi TJ_k - J_{k+1}||_D^2$
\ENDWHILE
\end{algorithmic}
Notice that by definition of the projection operator $\Pi$ we can write $\Pi TJ_k = \Phi r_{k+1}$ where $$ r_{k+1} = \argmin_r ||TJ_k - \Phi r||_D^2$$ where $TJ_k(i) = \mathbf{E}_\mu[g(i,j) + J_k(j)]$ is the one step Bellman Error for $k^{th}$ estimate of $J$ and \\ $J_k = \Phi x_k$ for some $x_k \in \mathbf{R}^k$

To compute $r_{k+1}$, a gradient descent would result in the following update $$r := r + \beta (g_{ij} + \phi(j)^Tx_k - \phi(i)^Tr)\phi(i)$$

With this updated value of $r_{k+1}$ we solve the next minimation problem $\min_{J_k} ||\Phi r_{k+1} - J_k||_D^2 $
$$= \min_{x_k} ||\Phi r_{k+1} - \Phi x_k||_D^2$$

A gradient descent would result in the following update$$x_k := x_k - \gamma(\phi(i)^Tr_{k+1} - \phi(i)^Tx_k)\phi(i)$$

The resulting algorithm thus becomes,
\begin{algorithm}
\caption{Coordinate Descent TD(0)}
\hspace*{\algorithmicindent} \textbf{Input} A trajectory of states simulated according to policy $\mu$, $r_0, x_0$\\
\hspace*{\algorithmicindent} \textbf{Output} $J^* = \Phi r^* = \Phi x^*$
\begin{algorithmic}
\FOR{each state transition $(i,j)$ with reward $g_{ij}$}
\WHILE{$\delta$ tolerance}
\STATE compute $r := r + \beta (g_{ij} + \phi(j)^Tx_k - \phi(i)^Tr)\phi(i)$
\ENDWHILE
\STATE $r_{k+1} \leftarrow r$ 
\WHILE{$\delta$ tolerance}
\STATE  $x := x - \gamma(\phi(i)^Tr_{k+1} - \phi(i)^Tx)\phi(i)$
\ENDWHILE
\STATE $x_{k+1} \leftarrow x$ 
\ENDFOR
\end{algorithmic}
\end{algorithm}

\subsection{A sketch of convergence analysis}

Let us say that we have an $x_k$ at the begining of some iteration. 
Let us look at the first update equation :
$$r := r + \beta (g_{ij} + \phi(j)^Tx_k - \phi(i)^Tr)\phi(i)$$

This algorithm convergence $r \to r_{k+1}$ s.t $$\Pi T \Phi x_k = \Phi r_{k+1}$$

With this $r_{k+1}$ fixed we consider the next update equation $$x_k := x_k - \gamma(\phi(i)^Tr_{k+1} - \phi(i)^Tx_k)\phi(i)$$

This update converges $x_k \to r_{k+1}$ s.t. $x_{k+1} = r_{k+1}$

We use this value in the first update to get $r_{k+2}$ s.t $$\Pi T \Phi x_{k+1} = \Pi T \Phi r_{k+1} = \Phi r_{k+2}$$

Thus we see that running these two updates gives us a sequence $\{r_k)_{k>0}$ which satisfies the following updates 
$$\Pi T \Phi r_k = \Phi r_{k+1} \forall k = 1,2,..$$

This last equation is known to converge.

\subsection{Implementation Details and Extensions to Non-linear Function approximations and control problems}

For implementations sake we have used the following schema, where each gradient update step is performed sequentially, without waiting for $\delta$ convergence.

\begin{algorithm}
\caption{Alternating Coordinate Descent for TD(0)}
\hspace*{\algorithmicindent} \textbf{Input} A trajectory of states simulated according to policy $\mu$, $r_0, x_0$\\
\hspace*{\algorithmicindent} \textbf{Output} $J^* = \Phi r^* = \Phi x^*$
\begin{algorithmic}
\FOR{each state transition $(i,j)$ with reward $g_{ij}$}
\STATE $r_{k+1} = r_k + \beta (g_{ij} + + \phi(j)^Tx_k - \phi(i)^Tr_k)\phi(i)$
\STATE  $x_{k+1} = x_k - \gamma(\phi(i)^Tr_{k+1} - \phi(i)^Tx_k)\phi(i)$
\ENDFOR
\end{algorithmic}
\end{algorithm}

For non-linear function approximations, say $J(i) = \tilde{J}(i,r)$ where $r$ is a learnable parameter, we have used the following schema

\begin{algorithm}
\caption{Alternating Coordinate Descent for TD(0)}
\hspace*{\algorithmicindent} \textbf{Input} A trajectory of states simulated according to policy $\mu$, $r_0, x_0$\\
\hspace*{\algorithmicindent} \textbf{Output} $J^* = \tilde{J}(r^*) = \tilde{J}(x^*)$
\begin{algorithmic}
\FOR{each state transition $(i,j)$ with reward $g_{ij}$}
\STATE $r_{k+1} =$ one step update of $\min_{r_k}\{g_{ij} + \tilde{J}(j, x_k) - \tilde{J}(i, r_k)\}^2$ using your favourite optimizer.
\STATE  $x_{k+1} = $ one step update of $\min_{x_k}\{\tilde{J}(i,r_{k+1}) - \tilde{J}(i,x_k)\}^2$ using your favourite optimizer.
\ENDFOR
\end{algorithmic}
\end{algorithm}

This has suggested the following critic network design.

\subsubsection{Implementation as a deep neural network}
\label{network_archi}
Let us say we have the following transition $(i, g_{i,j}, j)$

We can implement this algorithm by the designing a cooperative neural network. It represents a cooperation in the sense that the output of each network reinforces the parameters of the other network.\par

\tikzset{every picture/.style={line width=0.75pt}} 

\begin{tikzpicture}[x=0.50pt,y=0.50pt,yscale=-1,xscale=1]

\draw   (109,84) -- (213,84) -- (213,124) -- (109,124) -- cycle ;
\draw    (64,106) -- (109,106) ;
\draw [shift={(111,106)}, rotate = 180] [color={rgb, 255:red, 0; green, 0; blue, 0 }  ][line width=0.75]    (10.93,-3.29) .. controls (6.95,-1.4) and (3.31,-0.3) .. (0,0) .. controls (3.31,0.3) and (6.95,1.4) .. (10.93,3.29)   ;
\draw   (14,106) .. controls (14,92.19) and (25.19,81) .. (39,81) .. controls (52.81,81) and (64,92.19) .. (64,106) .. controls (64,119.81) and (52.81,131) .. (39,131) .. controls (25.19,131) and (14,119.81) .. (14,106) -- cycle ;
\draw    (213,107) -- (263,107) ;
\draw [shift={(265,107)}, rotate = 180] [color={rgb, 255:red, 0; green, 0; blue, 0 }  ][line width=0.75]    (10.93,-3.29) .. controls (6.95,-1.4) and (3.31,-0.3) .. (0,0) .. controls (3.31,0.3) and (6.95,1.4) .. (10.93,3.29)   ;
\draw   (265,107) .. controls (265,93.19) and (276.19,82) .. (290,82) .. controls (303.81,82) and (315,93.19) .. (315,107) .. controls (315,120.81) and (303.81,132) .. (290,132) .. controls (276.19,132) and (265,120.81) .. (265,107) -- cycle ;
\draw   (475,87) -- (579,87) -- (579,127) -- (475,127) -- cycle ;
\draw    (475,109) -- (432,109) ;
\draw [shift={(430,109)}, rotate = 360] [color={rgb, 255:red, 0; green, 0; blue, 0 }  ][line width=0.75]    (10.93,-3.29) .. controls (6.95,-1.4) and (3.31,-0.3) .. (0,0) .. controls (3.31,0.3) and (6.95,1.4) .. (10.93,3.29)   ;
\draw   (380,109) .. controls (380,95.19) and (391.19,84) .. (405,84) .. controls (418.81,84) and (430,95.19) .. (430,109) .. controls (430,122.81) and (418.81,134) .. (405,134) .. controls (391.19,134) and (380,122.81) .. (380,109) -- cycle ;
\draw    (627,106) -- (582,105.04) ;
\draw [shift={(580,105)}, rotate = 361.22] [color={rgb, 255:red, 0; green, 0; blue, 0 }  ][line width=0.75]    (10.93,-3.29) .. controls (6.95,-1.4) and (3.31,-0.3) .. (0,0) .. controls (3.31,0.3) and (6.95,1.4) .. (10.93,3.29)   ;
\draw   (627,106) .. controls (627,92.19) and (638.19,81) .. (652,81) .. controls (665.81,81) and (677,92.19) .. (677,106) .. controls (677,119.81) and (665.81,131) .. (652,131) .. controls (638.19,131) and (627,119.81) .. (627,106) -- cycle ;
\draw   (322,190) .. controls (322,176.19) and (333.19,165) .. (347,165) .. controls (360.81,165) and (372,176.19) .. (372,190) .. controls (372,203.81) and (360.81,215) .. (347,215) .. controls (333.19,215) and (322,203.81) .. (322,190) -- cycle ;
\draw   (260,98) .. controls (260,85.3) and (270.3,75) .. (283,75) -- (352,75) .. controls (364.7,75) and (375,85.3) .. (375,98) -- (375,200) .. controls (375,212.7) and (364.7,223) .. (352,223) -- (283,223) .. controls (270.3,223) and (260,212.7) .. (260,200) -- cycle ;
\draw   (243,76.8) .. controls (243,67.52) and (250.52,60) .. (259.8,60) -- (414.2,60) .. controls (423.48,60) and (431,67.52) .. (431,76.8) -- (431,127.2) .. controls (431,136.48) and (423.48,144) .. (414.2,144) -- (259.8,144) .. controls (250.52,144) and (243,136.48) .. (243,127.2) -- cycle ;

\draw (24,98.4) node [anchor=north west][inner sep=0.75pt]    {$\phi ( i)$};
\draw (278,95.4) node [anchor=north west][inner sep=0.75pt]    {$J_{r}( i)$};
\draw (154,93.4) node [anchor=north west][inner sep=0.75pt]    {$r$};
\draw (640,101.4) node [anchor=north west][inner sep=0.75pt]    {$\phi ( i)$};
\draw (394,98.4) node [anchor=north west][inner sep=0.75pt]    {$J_{x}( i)$};
\draw (518,97.4) node [anchor=north west][inner sep=0.75pt]    {$x$};
\draw (331,183.4) node [anchor=north west][inner sep=0.75pt]    {$TJ_{x}( i)$};

\end{tikzpicture}

Here $TJ_x(i) = g_{i,j} + J_x(j)$.

The box over $TJ_x(i)$ and $J_r(i)$ represents the error which is backpropagated over the left hand network.

The box over $J_r(i)$ and $J_x(i)$ represents the error that is backpropagated over the right hand side network.

\subsubsection{Control Problems using Q- factors}

For control problems we have the following Q-factor alternative algorithm, where each $Q(i,u) = \tilde{Q}(i,u,r)$ where $r$ is a learnable parameter and $i \in \mathcal{X}$ and $u \in \mathcal{U}$

\begin{algorithm}
\caption{Alternating Coordinate Descent for Q-Learning}
\hspace*{\algorithmicindent} \textbf{Input} An initial state $i_0$ $r_0, x_0$\\
\hspace*{\algorithmicindent} \textbf{Output} $Q^* = \tilde{Q}(r^*) = \tilde{Q}(x^*)$
\begin{algorithmic}
\FOR{current state $i$}
\STATE choose $u$ as $\max_u \tilde{Q}(i,u,r_k)$ with probability $\epsilon$ or choose a random action
\STATE Based on action $u$ observe next state $j$ and reward $g_{ij}$
\STATE $r_{k+1} =$ one step update of $\min_{r_k}\{g_{ij} + \max_v\tilde{Q}(j,v, x_k) - \tilde{Q}(i,u, r_k)\}^2$ using your favourite optimizer.
\STATE  $x_{k+1} = $ one step update of $\min_{x_k}\{\tilde{Q}(i,u,r_{k+1}) - \tilde{Q}(i,u, x_k)\}^2$ using your favourite optimizer.
\STATE update current state as $j$
\ENDFOR
\end{algorithmic}
\end{algorithm}

\subsubsection{Control Problems using Actor-Critic}

Since our algorithm is itself a critic network in essence, it can be used with any other actor network in conjunction to solve control problems.

We shall see more of these in the experimant section.

\section{A simple example}

Let us consider the following problem. A Markov Chain that has 3 states $A$, $B$ and $C$. The transition probabilities are as marked.\par

\tikzset{every picture/.style={line width=0.75pt}} 

\begin{tikzpicture}[x=0.75pt,y=0.75pt,yscale=-1,xscale=1]

\draw   (49,116) .. controls (49,102.19) and (60.19,91) .. (74,91) .. controls (87.81,91) and (99,102.19) .. (99,116) .. controls (99,129.81) and (87.81,141) .. (74,141) .. controls (60.19,141) and (49,129.81) .. (49,116) -- cycle ;
\draw   (176,49) .. controls (176,35.19) and (187.19,24) .. (201,24) .. controls (214.81,24) and (226,35.19) .. (226,49) .. controls (226,62.81) and (214.81,74) .. (201,74) .. controls (187.19,74) and (176,62.81) .. (176,49) -- cycle ;
\draw   (185,187) .. controls (185,173.19) and (196.19,162) .. (210,162) .. controls (223.81,162) and (235,173.19) .. (235,187) .. controls (235,200.81) and (223.81,212) .. (210,212) .. controls (196.19,212) and (185,200.81) .. (185,187) -- cycle ;
\draw    (95,128) -- (183.33,185.9) ;
\draw [shift={(185,187)}, rotate = 213.25] [color={rgb, 255:red, 0; green, 0; blue, 0 }  ][line width=0.75]    (10.93,-3.29) .. controls (6.95,-1.4) and (3.31,-0.3) .. (0,0) .. controls (3.31,0.3) and (6.95,1.4) .. (10.93,3.29)   ;
\draw    (99,109) -- (176.44,47.25) ;
\draw [shift={(178,46)}, rotate = 501.43] [color={rgb, 255:red, 0; green, 0; blue, 0 }  ][line width=0.75]    (10.93,-3.29) .. controls (6.95,-1.4) and (3.31,-0.3) .. (0,0) .. controls (3.31,0.3) and (6.95,1.4) .. (10.93,3.29)   ;
\draw    (189,71) .. controls (193.9,106.28) and (204.56,104.1) .. (210.63,74.83) ;
\draw [shift={(211,73)}, rotate = 460.95] [color={rgb, 255:red, 0; green, 0; blue, 0 }  ][line width=0.75]    (10.93,-3.29) .. controls (6.95,-1.4) and (3.31,-0.3) .. (0,0) .. controls (3.31,0.3) and (6.95,1.4) .. (10.93,3.29)   ;
\draw    (201,163) .. controls (211.67,121.29) and (219.52,144.5) .. (222.72,163.27) ;
\draw [shift={(223,165)}, rotate = 261.03] [color={rgb, 255:red, 0; green, 0; blue, 0 }  ][line width=0.75]    (10.93,-3.29) .. controls (6.95,-1.4) and (3.31,-0.3) .. (0,0) .. controls (3.31,0.3) and (6.95,1.4) .. (10.93,3.29)   ;

\draw (67,105) node [anchor=north west][inner sep=0.75pt]   [align=left] {A};
\draw (194,41) node [anchor=north west][inner sep=0.75pt]   [align=left] {B};
\draw (202,177) node [anchor=north west][inner sep=0.75pt]   [align=left] {C};
\draw (121,49.4) node [anchor=north west][inner sep=0.75pt]    {$\frac{1}{2}$};
\draw (124,161.4) node [anchor=north west][inner sep=0.75pt]    {$\frac{1}{2}$};
\draw (210,88.4) node [anchor=north west][inner sep=0.75pt]    {$1$};
\draw (224,132.4) node [anchor=north west][inner sep=0.75pt]    {$1$};

\end{tikzpicture}

The each transiotion has $0$ reward except for the self transition at $B$ and $C$ which are $1$. We choose the invariant distribution of $\pi = [0,0.8,0.2]$ respectively for states $A, B$ and $C$.

We have a simple state-feature representation $\phi(A) = \epsilon$, $\phi(B) = -1$ and $\phi(C) = 1$ where $\epsilon$ is a small positive numbers to ensure we don't run into zero gradient estimates. 

We first plot the objective function $$||TJ_\theta-J_\theta||_D^2$$, which is used for gradient based algorithms w.r.t the tunable paramater $\theta$

\begin{figure} 
  \centering
  \begin{minipage}{0.35\textwidth}
    \centering
    \includegraphics[width = \linewidth]{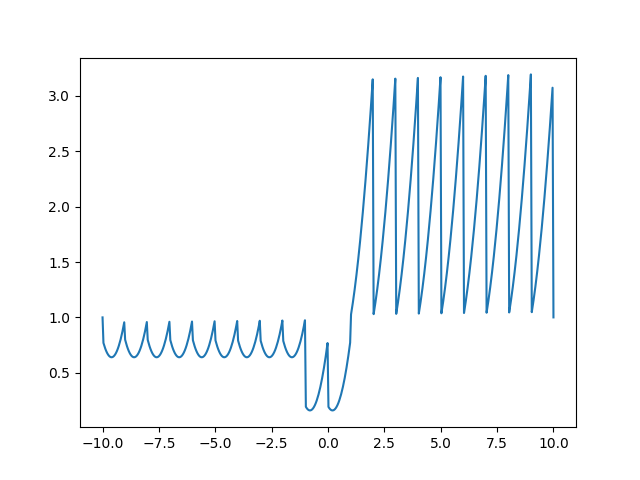}
    \caption{MSBE}
    \label{fig:msbe}
  \end{minipage}%
  \begin{minipage}{0.35\textwidth}
    \centering
    \includegraphics[width = \linewidth]{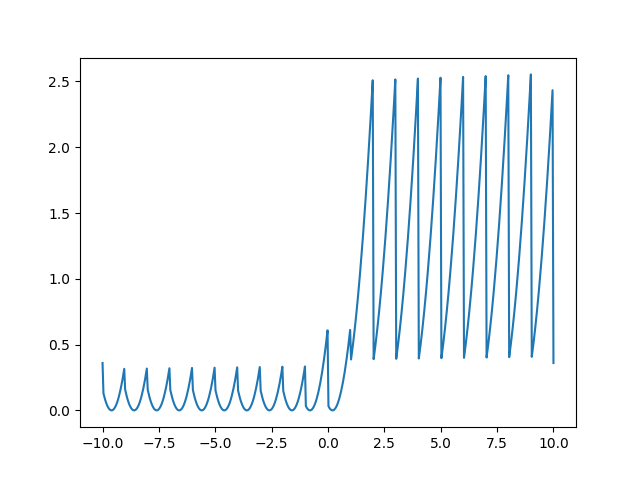}
    \caption{MSPBE}
    \label{fig:mspbe}
  \end{minipage}
  \begin{minipage}{0.35\textwidth}
    \centering
    \includegraphics[width = \linewidth]{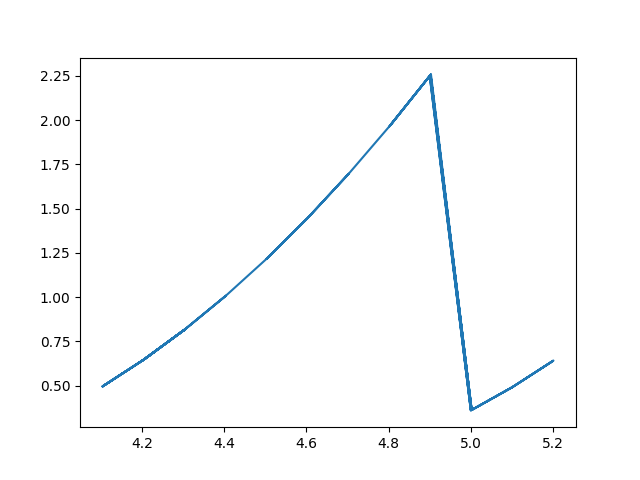}
    \caption{TD(0) with $\theta_0=5$}
    \label{fig:td0_positive}
  \end{minipage}%
  \begin{minipage}{0.35\textwidth}
    \centering
    \includegraphics[width = \linewidth]{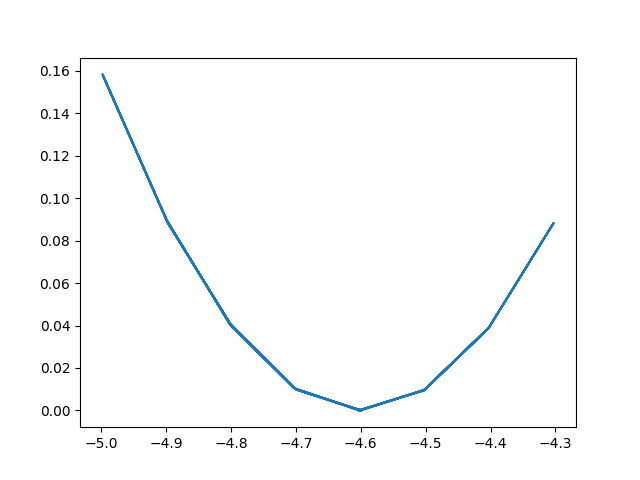}
    \caption{TD(0) $\theta_0=-5$}
    \label{fig:td0_negative}
  \end{minipage}
  \begin{minipage}{0.35\textwidth}
    \centering
    \includegraphics[width = \linewidth]{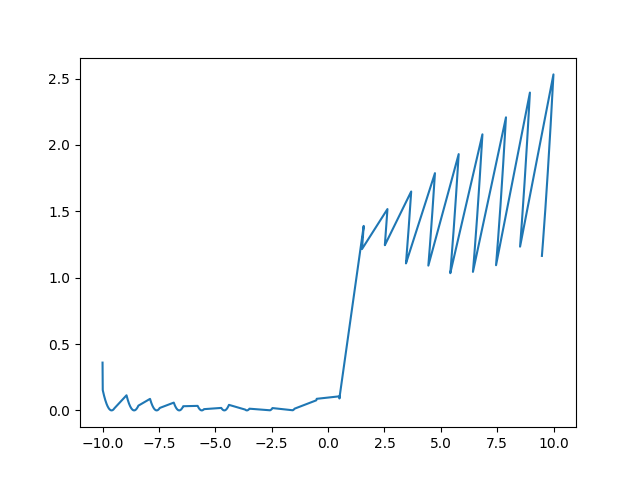} 
  \caption{MSPBE gradient \\based algorithm}
  \label{fig:mspbe_grad_based}
  \end{minipage}%
  \begin{minipage}{0.35\textwidth}
    \centering
    \includegraphics[width = \linewidth]{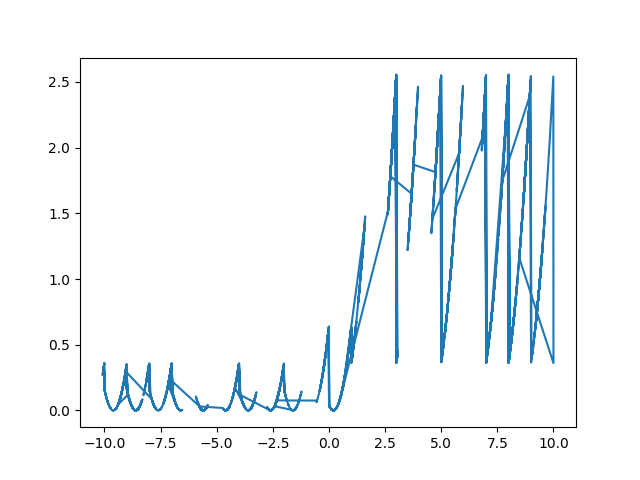} 
    \caption{MSPBE for \\GTD2 algorithm}
    \label{fig:mspbe_gtd2}
  \end{minipage}
  \begin{minipage}{0.35\textwidth}
    \centering
    \includegraphics[width = \linewidth]{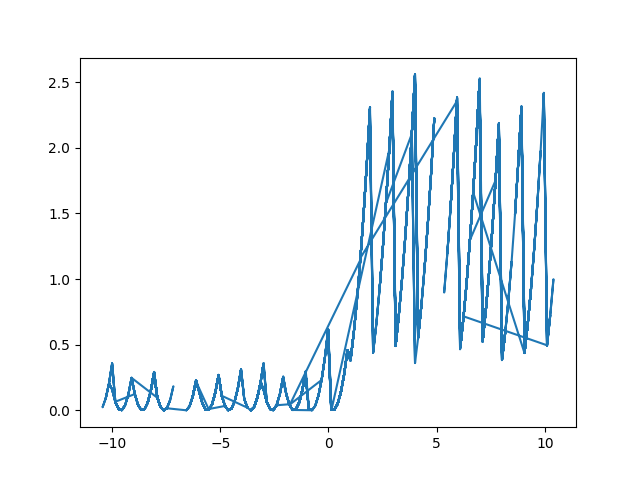} 
    \caption{MSPBE for THIS algorithm}
    \label{fig:mspbe_this_algo}
  \end{minipage}
\end{figure}

Let us compare this with the MSPBE objective function $$||\Pi TJ_\theta - J_\theta||_D^2$$

We see that the MSPBE error (Figure \ref{fig:mspbe}) is less than the MSBE (Figure \ref{fig:msbe}).

Let us now look at how the $TD(\lambda)$ performs. The $TD(\lambda)$ is known to track the MSPBE. The $TD(\lambda)$ algorithm also has the fastest convergence rates compared to gradient based methods.

We plot the $MSPBE$ of the $TD(0)$ algorithm with the updates of the tunable parameter $\theta$.

We first show the result when the initial point of $\theta_0 = 5.0$ (Figure \ref{fig:td0_positive}) From the $MSPBE$ graph we see that it would converge to a suboptimal solution of $0.5$

The learning rate has been kept fixed at $0.001$, instead of using decreasing step size. This results in some noise but we ignore that, as the point is made clear that a local minima of $0.5$ is reached.

We are also confirmed that while start at initial values of $\theta_0 = -5.0$ we reach the global minima of $0$ (Figure \ref{fig:td0_negative}).

Again a bit of overshoot is visible because of the randomness in the simulation and using constant step sizes.

We now look at the gradient based algorithms which aim at tracking $$||TJ_\theta-J_\theta||_D^2$$. 

\begin{figure} 
  \centering
  \begin{subfigure}[b]{0.3\linewidth}
    \includegraphics[width=\linewidth]{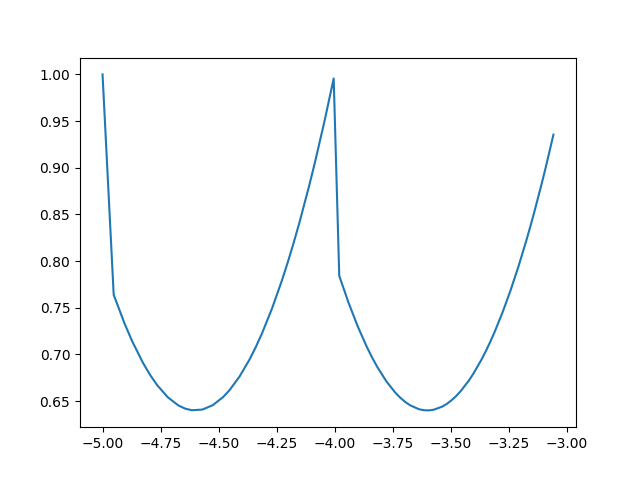}
     \caption{$\theta_0 = -5$}
  \end{subfigure}
  \begin{subfigure}[b]{0.3\linewidth}
    \includegraphics[width=\linewidth]{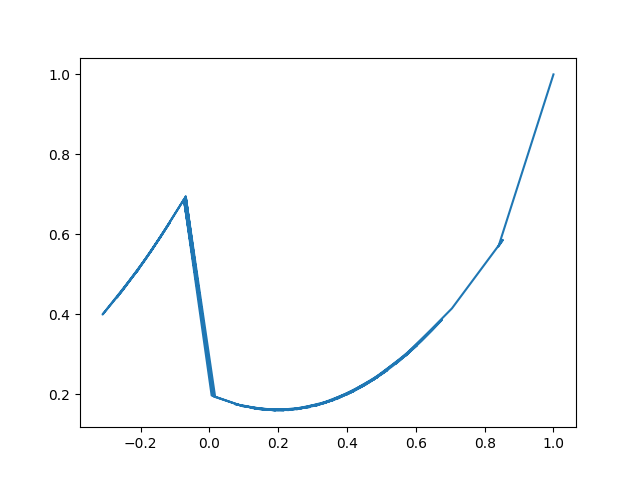}
    \caption{$\theta_0 = 1$}
  \end{subfigure}
  \begin{subfigure}[b]{0.3\linewidth}
    \includegraphics[width=\linewidth]{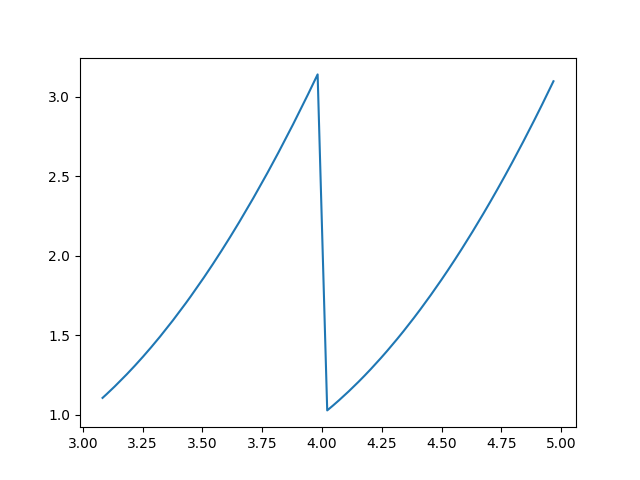}
    \caption{$\theta_0 = 5$}
  \end{subfigure}
  \caption{Gradient Based Algorithms}
  \label{fig:grad_based}
\end{figure}

We plot $||TJ_\theta-J_\theta||_D^2$ vs $\theta$.
The results are also confirmed by looking at $MSBE$ plot. We start with 3 different starting positions of $\theta_0$ : $-5, 1$ and $5$ (Figure \ref{fig:grad_based}).

That the gradient based algorithms are slower is confirmed by having taken learning rates $0.01$, $10$ times that of $TD$ algorithms.

The $MSBE$ optimal $\theta$ is also optimal for the $MSPBE$, however this may not be true as illustrated in Figure \ref{fig:mspbe_grad_based}. This figure tracks the $MSPBE$ with changing $\theta$ while using gradient based algorithms. We see that for negative values of $\theta$ the $MSPBE$ is $0$ which is consistent with the $MSPBE$ plot for negative values of $\theta$. However, for positive values of $\theta$ the local optima should be at $0.5$ whereas as shown in the figure, the gradient based algorithm tracks the $MSPBE$ at $1$.

Thus the gradient based algorithms do not track the $MSPBE$. Let us look at the GTD2 algorithm which does track the $MSPBE$ (Figure \ref{fig:mspbe_gtd2}).

However GTD2 has quite slow convergence. The leaning rate has been fixed to $100$ which is $10000$ times that of gradient based algorithms.

We now see that the proposed algorithm not only tracks the $MSPBE$ (Figure \ref{fig:mspbe_this_algo}) but the learning rate chosen is $0.1$, which is $1000$ times as less than GTD2 based algorithm. 

The derivation also shows that this can be implemented very easily since it is a gradient based algorithm by itself.

\section{Experiments}

To analyze the effects of the proposed network update rule we carried out the experiments in two different existing frameworks:\textit{Deep Q-Network(DQN)}\citep{mnih2015} and \textit{Deep Deterministic Policy Gradient(DDPG)}\citep{lillicrap2016}. We are using these existing frameworks because we aim to investigate the impact of proposed update rule in different existing settings\footnote{Implementation code in available at https://github.com/kavitawagh/RLProject.git}. DQN and DDPG are the benchmark algorithms in discrete action space and continuous action space setting, respectively.\par

In the following sections we explain how have we implemented the proposed update rule in DQN and DDPG framework and we also present the results of our experiments. 

\subsection{DQN}

For MDPs having discrete action space, DQN is the non-linear approximation of Q-function implemented using neural networks. The state $s$ is the input to Q-Network and for each action $a$ in the action space, Q-Network outputs the Q-value $Q(S, a)$. Greedy policy is used to select single Q-value out of all outputs of the Q-Network, where Q-value selected is $\max_{a} Q(s, a)$. \par

\begin{figure} 
\centering
\includegraphics[width=0.7\linewidth]{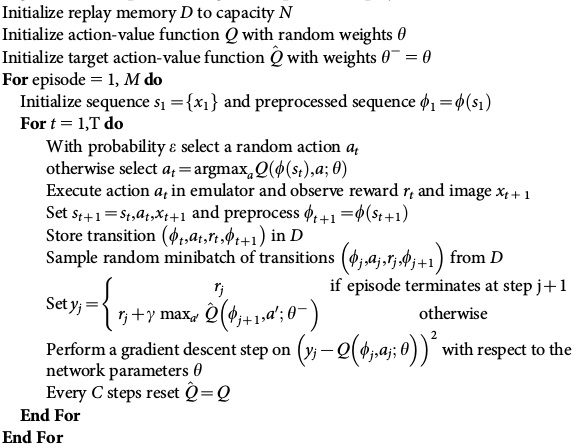} 
\caption{Original DQN Algorithm\citep{mnih2015}}
\label{fig:dqn_algo}
\end{figure}

Non-linear function approximation is known to be unstable or even diverge when used to approximate Q-value function. One of the reason for the instability is the correlation between Q-value $Q(s, a)$ and target Q-value $\{reward+\gamma*\max_{a'}Q(s', a')\}$ when same network is used to predict Q-values and target Q-values. Hence in DQN framework, target Q-values are predicted by different Q-Network whose weight parameters are updated periodically. DQN also uses a replay buffer for training the Q-Network with a batch of transitions. Every time an agent takes some action in environment, the state transition tuple $(current\_state, action, reward, next\_state)$ is stored in the replay buffer. In the network training step, a random batch of transition tuples is sampled from replay buffer and used in network optimization. Figure \ref{fig:dqn_algo} presents the original DQN algorithm.\par

\subsubsection{Modification in algorithm}
\begin{figure} 
\centering
\includegraphics[width=0.8\linewidth]{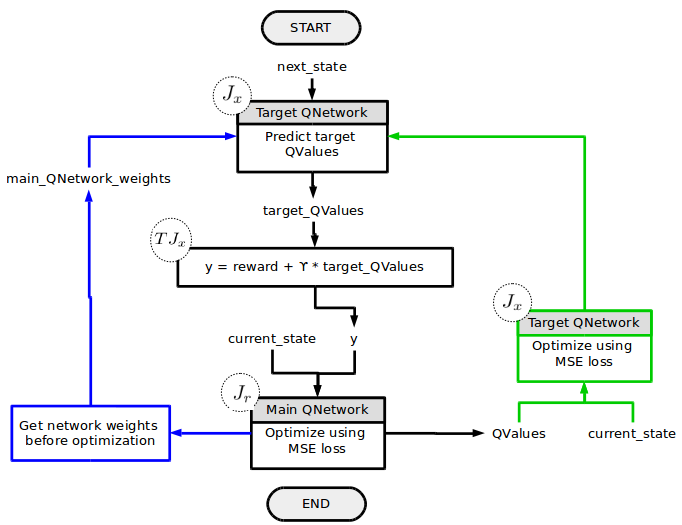} 
\caption{Implementation of proposed update rule in DQN framework}
\label{fig:newalgo_dqn}
\end{figure}

Figure \ref{fig:newalgo_dqn} shows pseudo-flowchart for DQN algorithm along with the proposed update rule. The blocks in black border indicate the processing steps that belongs to original DQN algorithm and we are using as it is in modified algorithm.  The shapes in blue indicate the processing steps in the original DQN algorithm those we have replaced with our processing steps highlighted with green color. The names in circles indicate the relation between our implementation and the network architecture proposed in Section \ref{network_archi}.\par

In every environment step, we sample a batch of transitions from the replay buffer. Feeding $next\_state$ to target Q-network we get $target\_QValue$. The Bellman operator is applied using $target\_QValue$ to get the target $y$ to optimize the main Q-network. Main Q-network is then optimized by minimizing the mean square error(MSE) loss between its prediction and $y$. In original DQN algorithm, before optimizing the main Q-network, the weights of target Q-network are set equal to the weights of main Q-network and then main Q-network is optimized. While in our modification, instead of just copying the weights, we take an optimization step on target Q-network.

\subsubsection{Environment}
We trained our Q-Network for OpenAI Gym CartPole environment. Cart-Pole is the classical reinforcement learning problem where the aim is to balance the pole upright on the cart by moving the cart left or right. State is represented by a list of observations [cart position, cart velocity, pole angle, pole velocity at the tip]. Actions include pushing the cart left or right. Reward is +1 for every step taken by the agent, including the termination step. Episode terminates if pole angle is more than $\pm12^{\circ}$ or the accumulated reward is equal to 200.  

\subsubsection{Results}
\begin{figure}
\centering
\begin{subfigure}{0.5\textwidth}
  \centering
  \includegraphics[width=\linewidth]{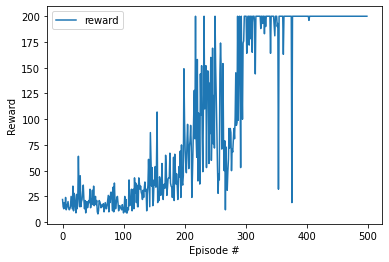}
  \caption{Original DQN algorithm}
  \label{fig:cartpole_dqn}
\end{subfigure}%
\begin{subfigure}{0.5\textwidth}
  \centering
  \includegraphics[width=\linewidth]{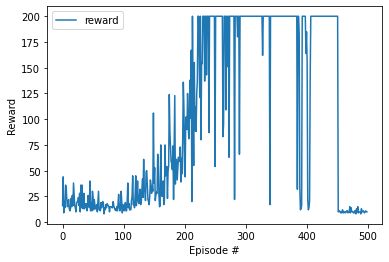}
  \caption{DQN algorithm with proposed update rule}
  \label{fig:cartpole_newalgo}
\end{subfigure}
\caption{Plot of training reward against episodes using DQN for CartPole environment}
\label{fig:cartpole}
\end{figure}

We trained the Q-Network using original DQN algorithm and DQN algorithm with proposed update rule. Figure \ref{fig:cartpole} shows the plot of actual training reward against the episode numbers. We can see that the results of proposed update rule are comparable to that of the original DQN algorithm. After achieving the highest, the reward goes to zero in some episodes. This is possible because of the stochastic nature of the problem and the behaviour policy being used. We are using epsilon-greedy policy where minimum epsilon value is 0.1, that is agent can take random action with probability 0.1.

\subsection{DDPG}

DDPG is actor-critic algorithm based on deterministic policy gradient for the problems having continuous action space. It uses the non-linear approximation of critic which models the Q-value function and actor which models the deterministic policy $\mu$. Critic network predicts the Q-value $Q(s, a)$ given the state $s$ and action $a$ as input. Actor network predicts the action $\mu(s)$ given state $s$ as the input. \par
\begin{figure} 
\centering
\includegraphics[width=0.8\linewidth]{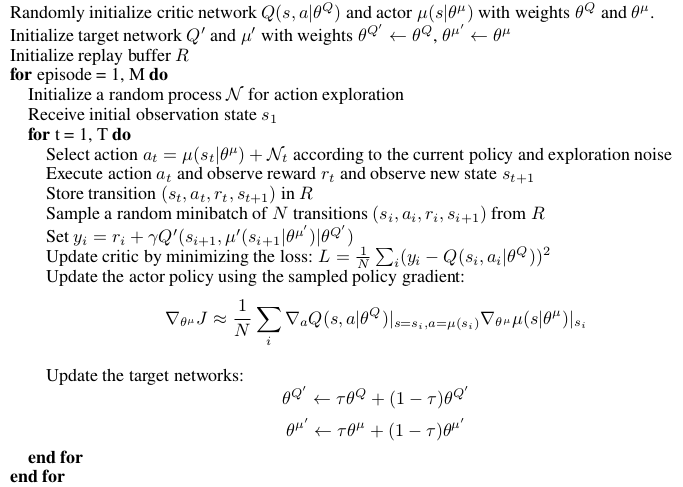} 
\caption{Original DDPG Algorithm\citep{lillicrap2016}}
\label{fig:ddpg_algo}
\end{figure}

Based on the current network parameters, actor network predict action. Using this predicted action, critic network predicts the Q-value. This Q-value is used as feedback for actor saying how good or bad the action predicted by actor is. The key concept that DDPG based on is the form of the deterministic policy gradient. The gradient of the state-value function $J(.)$ w.r.t to actor network parameter have a nice implementable form involving the gradient of Q-value function and the gradient of policy(actor) network w.r.t network parameters. Optimizing the actor network with the gradient-ascent step in the direction of policy gradient guarantees to asymptotically converge into optimal optimal policy. 
Formally, policy gradient is given by,
\[\nabla_{\theta}J = \nabla_aQ(s, a)|_{a=\mu(s)} \nabla_{\theta}\mu(s)\]
where $J$ is the state-value function,\\
$Q(.,.)$ is the Q-value function(critic network),\\
$\mu$ is the deterministic policy(actor network with parameters $\theta$)\\
Figure \ref{fig:ddpg_algo} presents the original DDPG algorithm.

\subsubsection{Modification in algorithm}
\begin{figure} 
\centering
\includegraphics[width=0.8\linewidth]{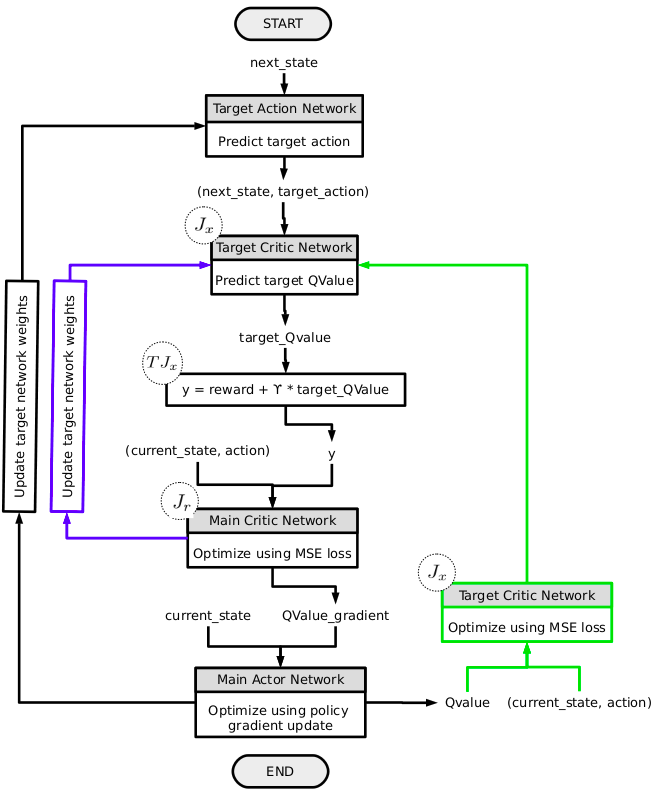} 
\caption{Implementation of proposed update rule in DDPG framework}
\label{fig:newalgo_ddpg}
\end{figure}

Figure \ref{fig:newalgo_ddpg} shows pseudo-flowchart for DDPG algorithm along with the proposed update rule. The blocks in black border indicate the processing steps that belongs to original DDPG algorithm and we are using as it is in modified algorithm.  The shapes in blue indicate the processing steps in the original DDPG algorithm those we have replaced with our processing steps highlighted with green color. The names in circles indicate the relation between our implementation and the network architecture proposed in Section \ref{network_archi}.\par

There are two actor networks: main actor and target actor; and two critic networks: main critic and target critic. The predictions of main networks are actually used to select action for the agent while target networks are used to calculate Bellman operator. First, a batch of transitions ($current\_state, action, reward, next\_state$) is sampled from the replay buffer. Feeding $next\_state$ as input to target actor, we get $target\_action$. This $target\_action$ with $next\_state$ in target critic gives $target\_QValue$. Then we calculate Bellman operator value $y$ which is used as target to optimize the main critic network. In original DDPG algorithm, the weights of the both target models are updated using soft update rule. In our modification, we replace the weight update step for target critic by an optimization step.

\subsubsection{Environment}
We trained our actor-critic network on OpenAI Gym BipedalWalker-v3 environment. BipedalWalker is a robot with a LIDAR system and two legs having four joints, two in each leg. The goal is to train the robot to walk as fast as possible on the simple terrain(with no obstacles). The environment state contains the LIDAR sensor measurements, joint positions, velocity, etc. The action is to apply the torques on four joints. The torque for any joint is value in [-1, 1] and this makes it continuous action space problem. The +1 reward is given for moving one step forward, total 300+ points up to the far end. If the robot falls, it gets -100 reward.

\subsubsection{Results}
\begin{figure}
\centering
\begin{subfigure}{0.3\textwidth}
  \centering
  \includegraphics[width=\linewidth]{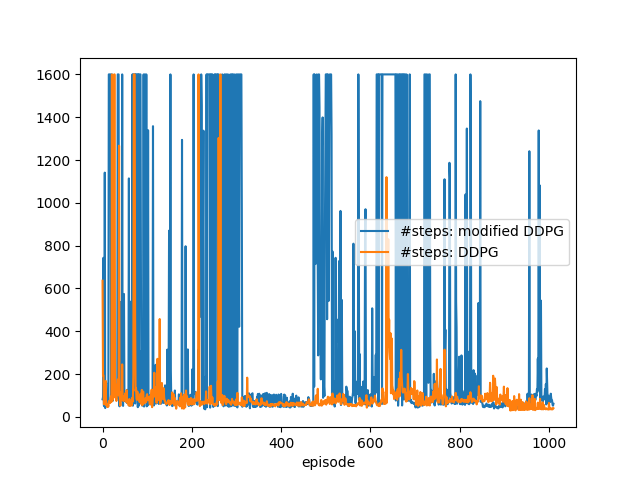}
  \caption{No. of steps before robot falls down}
  \label{fig:steps}
\end{subfigure}%
\begin{subfigure}{0.3\textwidth}
  \centering
  \includegraphics[width=\linewidth]{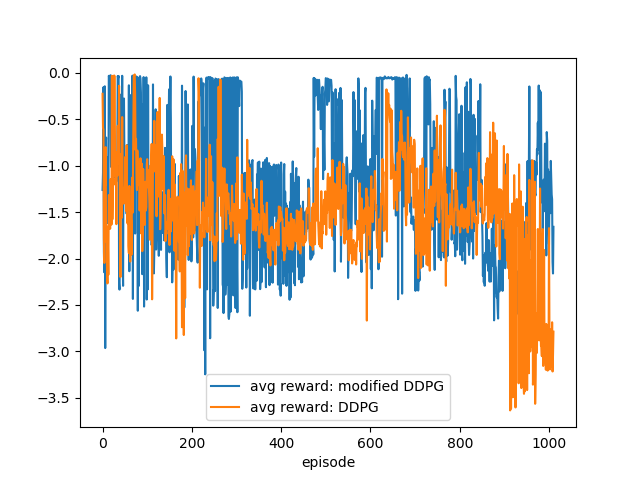}
  \caption{Avg reward per episode}
  \label{fig:avg_reward}
\end{subfigure}%
\begin{subfigure}{0.3\textwidth}
  \centering
  \includegraphics[width=\linewidth]{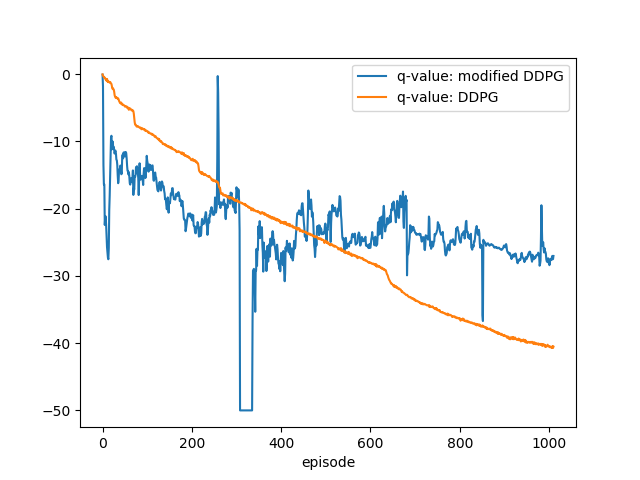}
  \caption{Q-Value}
  \label{fig:q_value}
\end{subfigure}

\begin{subfigure}{0.3\textwidth}
  \centering
  \includegraphics[width=\linewidth]{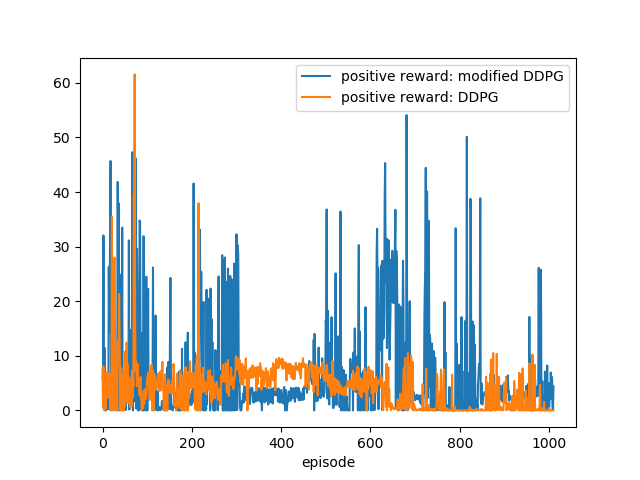}
  \caption{Per episode accumulated\\ positive reward}
  \label{fig:positive_reward}
\end{subfigure}%
\begin{subfigure}{0.3\textwidth}
  \centering
  \includegraphics[width=\linewidth]{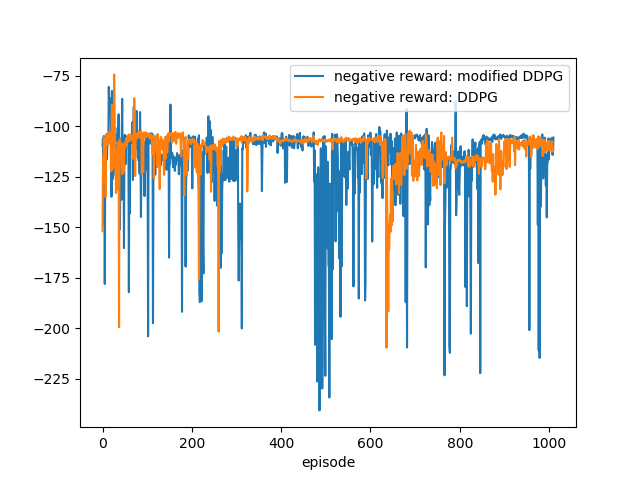}
  \caption{Per episode accumulated\\ negative reward}
  \label{fig:negative_reward}
\end{subfigure}%
\begin{subfigure}{0.3\textwidth}
  \centering
  \includegraphics[width=\linewidth]{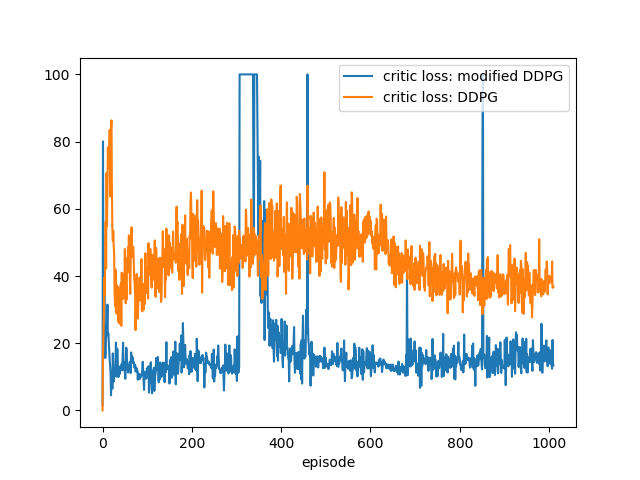}
  \caption{Critic loss}
  \label{fig:critic_loss}
\end{subfigure}

\caption{Training results for BipedalWalker environment using DDPG}
\label{fig:bipedal}
\end{figure}

We trained the actor-critic using original DDPG algorithm and DDPG algorithm with proposed update rule. Figure \ref{fig:bipedal} shows the plots for actual training. The no. of steps the robot is able to balance on its legs is larger in modified DDPG than that in the original DDPG (Figure \ref{fig:steps}). After visualizing the training episodes we came to know that agent in original DDPG is trying to move forward without taking the proper balanced pose first, and was falling immediately. While agent in modified DDPG first tries to take a balanced pose before trying to move forward. Since no. of steps the agent is alive is different for every episode we show the average reward per episode in Figure \ref{fig:avg_reward}. This figure indicates that the modified DDPG explores more than the original DDPG. Figure \ref{fig:q_value} shows the Q-value plot for both the algorithms.\par

The positive reward(Figure \ref{fig:positive_reward}) and negative reward(Figure \ref{fig:negative_reward}) both are higher in magnitude for modified DDPG. This is because we are plotting the total positive and negative rewards accumulated in an episode and since the modified DDPG survives for more no. of time steps it is obvious that it accumulates more reward in both positive and negative direction. In Figure \ref{fig:critic_loss} the critic loss in modified DDPG is always smaller than critic loss in original DDPG.

\section{Conclusion}
In our project, we came up with the new update rule for critic networks, in which we use and optimize two critic networks. We implemented the proposed update with DQN and DDPG algorithm on CartPole and BipedalWalker environments, respectively. Results of experiments showed that proposed update gives comparable results for both the algorithms. In future we aim to implement proposed update rule for other RL algorithms like TRPO, proximal policy, etc and analyze its performance. We also aim to prove rigorously the convergence of proposed update rule. 

\medskip
\small
\bibliography{references}

\end{document}